# Optimized Task Assignment and Predictive Maintenance for Industrial Machines using Markov Decision Process

Ali Nasir[1,5,*], Samir Mekid[2,5], Zaid Sawlan[3,5], and Omar Alsawafy[4,5]

[1]Control and Instrumentation Engineering Department, KFUPM, Dhahran, Saudi Arabia
[2]Mechanical Engineering Department, KFUPM, Dhahran, Saudi Arabia
[3]Mathematics Department, KFUPM, Dhahran, Saudi Arabia
[4]Systems Engineering Department, Dhahran, Saudi Arabia
[5]Interdisciplinary Research Center for Intelligent Manufacturing and Robotics, KFUPM, Dhahran, Saudi Arabia
[*]ali.nasir@kfupm.edu.sa

Abstract: This paper considers a distributed decision-making approach for manufacturing task assignment and condition-based machine health maintenance. Our approach considers information sharing between the task assignment and health management decision-making agents. We propose the design of the decision-making agents based on Markov decision processes. The key advantage of using a Markov decision process-based approach is the incorporation of uncertainty involved in the decision-making process. The paper provides detailed mathematical models along with the associated practical execution strategy. In order to demonstrate the effectiveness and practical applicability of our proposed approach, we have included a detailed numerical case study that is based on open source milling machine tool degradation data. Our case study indicates that the proposed approach offers flexibility in terms of the selection of cost parameters and it allows for offline computation and analysis of the decision-making policy. These features create and opportunity for the future work on learning of the cost parameters associated with our proposed model using artificial intelligence.

## 1 Introduction

The importance of predictive maintenance is well-recognized in the industrial sector for several reasons, e.g., it allows for the reduction in machine downtime, it helps in reducing the production cost, and it is useful in enhancing the life of machines. Consequently, predictive maintenance is one of the key areas of research among the scientific community. Initially, the predictive maintenance used to be time-based but later on (with the advances in sensing technology), condition-based maintenance (CBM) gained more popularity. Maintenance of machine tools involve two key stages, i.e., diagnosis and prognosis. Prognosis deals with the prediction of remaining useful life (RUL) of the machine whereas diagnosis is concerned with detection and identification of various faults in the machine. Major approaches for prognosis include data-based approaches, knowledge-based approaches, and physics (model) based approaches. Diagnosis on the other hand is based on centralized or distributed approaches [1]. Key challenges in predictive maintenance include 1) Dealing with the noisy sensor data, 2) Uncertainty in the operating conditions, and 3) Diversity of tasks assigned to the machine.

A comparison between time-based and condition-based maintenance strategies has been presented in [2]. Here, the authors have pointed out that although the condition-based maintenance is more beneficial from the practical point of view, still there are a lot of challenges related to the condition-based maintenance that require attention, e.g., development of robust sensing technologies, development of advanced decision-making software, and consolidation of associated theoretical results. Another survey [3] presents various CBM techniques and discusses the applications of CBM in aviation, automotive, defense, and other industries. A more recent review discusses the CBM optimization models for stochastically deteriorating systems [4]. In terms of dealing with uncertainty, there have been attempts to develop Markov decision process (MDP) based frameworks along with Brownian motion and Gamma process to represent deterioration chances. Data monitoring and analysis is also an important aspect of prognosis and condition monitoring as highlighted in [5]. The trends and popularity of various areas within condition-based maintenance for the past two decades have been presented in [6]. According to the results presented therein, the most popular aspects of maintenance are centralized versus distributed approaches and development of frameworks for prognosis of machines. A study in [7] highlights 30 different university-based research groups (across the globe) that are working on developing and improving frameworks for predictive maintenance.

Artificial intelligence-based methods have gained high popularity in recent times among the researcher in general and specifically among the community working on predictive maintenance. In this regard, a real-time machine tool failure forecast approach has been discussed in [8] where a data treatment system has been developed that automates the prediction and detection of failures. The treated (or preprocessed) data is compatible with the learning frameworks and can be used to develop decision-making strategy. Another similar attempt is presented in [9] where a reconfigurable prognostics platform has been developed that is capable of converting the sensor data into performance-related features. A deep learning-based approach for intelligent prognosis of machine tools has been discussed in [10]. In this work, an attention mechanism has been introduced to enhance the RUL prediction capability of long short-term memory (LSTM) network. A more specific approach for RUL prediction of aircraft engines has been presented in [11]. Here, deep bidirectional recurrent neural networks have been used to extract hidden features from the sensory data. RUL prediction has also been discussed in [12] where a novel scalarization method for raw sensory data has been developed based on the Weibull distribution.

Markov decision process is a tool that is used by at least three scientific communities, i.e., operations research community, computer science community, and control systems community. It is one of the very few frameworks that allow for optimal decision-making in the presence of uncertainty. The prognostics and predictive maintenance community has also developed MDP-based approaches. For example, in [13], LSTM and MDP have been used to develop a health status prediction-based approach for predictive maintenance. The weakness in this approach is that the operating conditions have not been accounted for. Another semi-Markov decision process-based approach for RUL prediction has been discussed in [14] where the variations in the operating conditions have also been accounted for in the model. A continuous-time Markov chain-based approach for RUL prediction has been discussed in [15] where a new aspect, i.e., task-oriented RUL prediction has been introduced. The term "task" has been devised to represent the user requirements regarding the availability and the performance of the machine(s).

Based on the above discussion, it is clear that there has been a substantial work done in the field of predictive maintenance. However, there is still a substantial research gap available in terms of combining user-based performance requirements, enhancement of lifetime of machine, and efficient predictive maintenance under variable operating conditions into a single performance management framework. Consequently, this paper presents an MDP-based framework that addresses this research gap and provides a consolidated MDP model that can be solved using stochastic dynamic programming to calculate an optimal decision-making policy for predictive maintenance and task management.

Continue later…

## 2 Background and Problem Formulation

Before we layout the problem statement and key objectives, we revisit the key concepts regarding the predictive maintenance and Markov decision process.

### 2.1 Predictive Maintenance

The details of predictive maintenance along with the description of classical methods can be found in [16]. Here, we layout the basic process for the condition-based maintenance and prognostics as shown in Figure *1*. Note that there are six major steps involved in the process. The process begins with the collection of real-time data from the machine sensors (e.g., current, temperature, vibrations, voltage, etc.). Then this data is pre-processed for removing the sensor noise. The resulting noise-free sensor data is analyzed for the extraction of features. A feature is based on certain properties of the data, e.g., mean, variance, max, min etc. Definition of features itself is a complex task but it suffices here to understand that a feature is an indicator of the health and/or performance status of the machine. Based on the extracted features, fault classification is performed which is a key step. Note that fault classification does include no-fault scenario. Next step is to predict the evolution of any faults detected. The prediction step is also used to determine the remaining useful life of the machine and the chances of occurrence of any faults in future. Based on the predictions, the maintenance of the machine is scheduled (or rescheduled).

The key steps in predictive maintenance are fault classification and prediction of fault evolution. There are multiple ways to execute these steps. For example, one could use the comparison between the expected and actual behavior of the system where the prediction of the system behavior is obtained from the mathematical model. Another approach is to use data-based training of the neural networks and utilization of trained networks for fault classification and prediction of the fault evolution. While the mathematical modeling offers the tradeoff between complexity and accuracy, a similar problem also exists for the neural networks-based approach where the tradeoff is between the amount of data used for training and the accuracy of the prediction (accurate prediction requires more training and more data). Due to these challenges, there is always some uncertainty involved in the process of fault classification and prediction of fault evolution. We intend to address this issue by proposing a Markov decision-based approach where the uncertainty is modeled in the optimal decision-making problem.

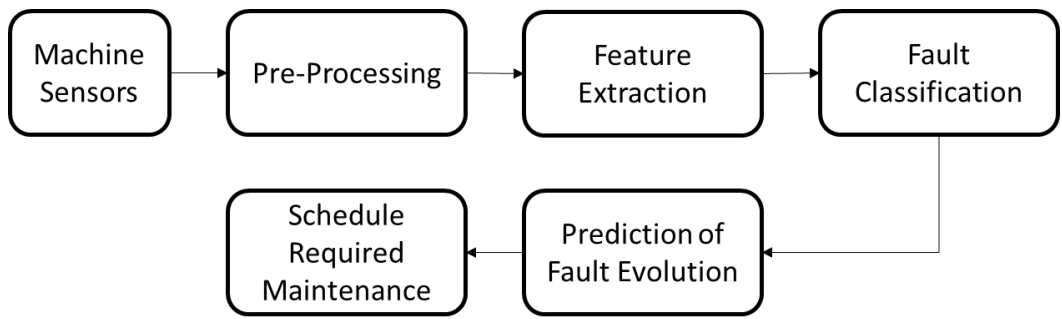

Figure 1: The process of predictive maintenance [16]

### 2.2 Markov Decision Process

Markov decision process is one of the most widely used framework for stochastic decision making [17]. A Markov decision process is composed of five main elements, i.e., a set of states ($S$), a set of decisions ($D$), a reward (or cost) function that depends upon the states and the decisions ($R$), a state transition probability tensor (or function) that contains (or describes) the probabilities of transition from one state to another for each possible decision option ($T$), and a discount factor which is a number ranging between zero and one and it depicts the depreciation on the value of future rewards (or cost) with each passing decision epoch ($\gamma$). A decision epoch in an MDP is time interval between the two consecutive decisions. It is a common misconception that the decision epoch is a fixed amount of time whereas in fact the decision-making based on MDP can be event-driven where time interval between two consecutive events may not be a constant. One may represent an MDP model as a five-tuple as follows

$$MDP \leftarrow \{S, D, R, T, \gamma\} \quad (1)$$

Once a stochastic decision-making problem is formulated as an MDP model, one can use a stochastic dynamic programming algorithm such as value iteration in order to solve the MDP model for an optimal decision-making policy. The value iteration algorithm iteratively solves the following equation

$$V^{t+1}(s) = \max_{d \in D} \left\{ R(s,d) + \sum_{s' \in S} \gamma T(s', d, s) V^t(s') \right\} \quad (2)$$

Here $V^t(s)$ is the value of a state $s$ at the decision epoch $t$. $R(s,d)$ is a state and decision dependent reward function. Here we assume that $R(s,d) > 0, \forall (s \in S, d \in D)$. $T(s', d, s)$ represents the probability of transitioning from state $s$ to the state $s'$ after opting for the decision $d$. It is well-known that the values of all the states converge to the optimal ones asymptotically for any nonnegative initial choice of values. The speed of convergence depends on the value of the discount factor $\gamma$. The convergence is slow if $\gamma$ is close to one and is fast if the value of $\gamma$ is close to zero. Once the state values converge, the optimal state values $V^*(s) \forall s \in S$ are used to calculate the optimal decision-making policy $\pi^*(s): S \rightarrow D$ using the following equation

$$\pi^*(s) = arg\max_{d \in D} \left\{ R(s,d) + \sum_{s' \in S} \gamma T(s', d, s) V^*(s') \right\} \quad (3)$$

Note that the optimal policy provides a complete decision-making solution as it yields the best decision for any given state. A major advantage of using MDP-based decision-making is that the optimal policy is computed offline which saves precious computation time during the machine operation.

## 2.3 Problem Description

Based on the preceding discussion and background information, we address the following problem in this paper

*Develop a decision-making framework for predictive maintenance that has the ability to reason about changing operating conditions, best possible machine performance, and task preferences set by the user. The decision-making shall be optimized with respect to the assignment of user-defined tasks to appropriate machines in such a way that the performance requirements are met, the life of machines is maximized, and the downtime is minimized.*

Solving the above-mentioned problem involves the following tasks. 1) Identification of appropriate state variables that provide adequate amount of information that is required for associated decision-making. 2) Identification of possible decisions in each state. 3) Appropriate definition of the reward function that reflects the objectives of the problem accurately. 4) Identification of random variables in the problem and their associated conditional dependencies that can be used to determine state transition probabilities. 5) Demonstration of the solution and real-life implementation of the proposed framework indicating satisfactory performance and achievement of the objectives.

An overview of the proposed solution for above-mentioned problem is shown in Figure *2*. Note that the decision-making is performed at two points. First, the MDP-based optimal policy determines the maintenance requirements and health status of the machines based on the data from the machine health sensors and the operational condition sensors. Second, the user-defined tasks with associated priority and performance requirements are assigned to appropriate machine(s) using an optimized machine selection policy. To the best of our knowledge, there is no existing solution that addresses the problem discussed in this paper. However, there is a lot of closely related work as discussed in the introduction section.

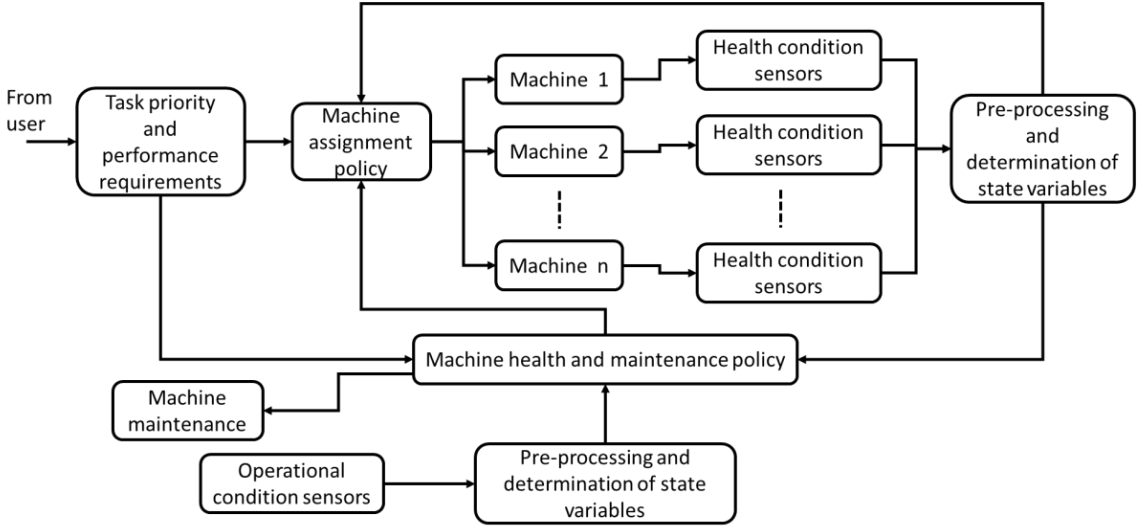

Figure 2: Overview of the proposed solution

# 3 Mathematical Model

In this section, we develop the mathematical model for the two types of decision-making blocks discussed in the previous section, i.e., machine health and maintenance decisions and task assignment decisions.

## 3.1 Machine Health and Maintenance Decision-making Model

In this section, we develop a Markov decision process-based model for determination of the machine RUL (we shall be associating machine RUL with the performance capability of the machine) and machine maintenance requirements. Specifically, we define the associated state variables, available decisions, state transition probabilities, and the state-dependent reward function.

### 3.1.1 State variables

In the existing literature, MDP-based decision-making regarding the RUL of machines [14] include two types of variables, i.e., discrete set of possible operating conditions ($E = \{1,2,...,M\}$) and a discrete set of machine health conditions ($\Lambda = \{1,2,...,L,F\}$). Note that the machine health is divided into $L$ working states and one failure state represented by $F$. Our formulation of the state space involves two changes, i.e., extension of the machine operating and health conditions to a multi-machine setup and addition of user-defined task flags. The advantage of having task flags is that the maintenance policy is able to understand the consequences of initiating maintenance for the machines in terms of possible downtime. The mathematical formulation of the set of states is given as

$$S = \{s_1, s_2, ..., s_N\},$$

$$s_i = \{e_{1,i}, e_{2,i}, \ldots, e_{n,i}, \lambda_{1,i}, \lambda_{2,i}, \ldots, \lambda_{n,i}, \tau_{1,i}, \tau_{2,i}, \ldots, \tau_{m,i}\},$$
$$e_{j,i} \in E = \{1, 2, \ldots, M\},$$
$$\lambda_{j,i} \in \Lambda = \{1, 2, \ldots, L, F\},$$
$$\tau_{j,i} \in T = \{0, 1, 2, 3\}. \tag{4}$$

Note that the above formulation assumes $n$ identical and independent machines where each machine can experience its own operating condition. The operating condition of $j^{th}$ machine in $i^{th}$ state is represented by $e_{j,i}$. We have assumed $M$ possible discrete operating conditions. Here we make an important assumption regarding the operating conditions, i.e., we assume that the operating conditions are ordered in terms of favorability for the machine (e.g., the most favorable operating condition is represented by 1 and least favorable operating condition is represented by $M$). The health condition of $j^{th}$ machine in $i^{th}$ state is represented by $\lambda_{j,i}$. The health condition variable may assume any one of the $L + 1$ possible values where $L$ values indicate degradation in the machine while it is still usable and the last value indicates the failure of the machine. We make similar assumption about the health condition as we did for the operating conditions, i.e., the health conditions are ordered in terms of severity of degradation (e.g., perfectly health machine is represented by 1 and least healthy machine is represented by $L$). Finally, we have assumed $m$ task variables in the state space where the $j^{th}$ task variable in $i^{th}$ state is represented by $\tau_{j,i}$ and it can take four possible values. The zero value for the task variable $\tau_{j,i}$ indicates that the $j^{th}$ task in $i^{th}$ state is inactive whereas the nonzero values of $\tau_{j,i}$ indicate the active status of the task with low ($\tau_{j,i} = 1$), medium ($\tau_{j,i} = 2$), and high ($\tau_{j,i} = 3$) priority. As per the state space formulation defined here, the total number of states in the MDP is given by $N = M^n \times (L + 1)^n \times 4^m$.

### 3.1.2 Set of Decisions

The set of decisions in our formulation includes initiating and scheduling maintenance for the available machines. Each decision is compound in a sense that it includes instructions regarding every machine. The maintenance is assumed to be of three types, i.e., 1) non-intrusive maintenance (such as oiling of the machine) where the normal operation is not disturbed, 2) Partially intrusive maintenance where the machine operation is affected but there is no need to shutdown the machine (for example, tightening of the bolts of non-moving parts of the machine), 3) Fully intrusive maintenance where the machine operation must be stopped for the maintenance to take place (e.g., replacement or service of the moving parts of the machine). Consequently, the mathematical representation of the decision $D$ is given by

$$D = \{d_1, d_2, \ldots, d_n\},$$
$$d_k \in \{0, 1, 2, 3\}. \tag{5}$$

Note that each decision is composed on $n$ flags where each flag $d_k$ can assume any of the four possible values where $d_k = 0$ means that no maintenance is requested for the $k^{th}$ machine; $d_k = 1$ means that non-intrusive maintenance is requested for the $k^{th}$ machine; $d_k = 2$ means that partially-intrusive maintenance is requested for the $k^{th}$ machine; and $d_k = 3$ means that fully-intrusive maintenance is requested for the $k^{th}$ machine. According to our formulation, there are $4^n$ possible decisions in each state.

### 3.1.3 Reward Function Formulation

The reward function for the decision-making process is formulated based on the objectives of the problem. Recall that in terms of the maintenance of the machines, the two main objectives are, maximization of the machine life and minimization of the down time. Consequently, the reward function is formulated as

$$R(s_i, D) = r_1 \left( \sum_{j=1}^{n} e_{j,i} \right) \left( \exp\left\{ -r_2 \sum_{j=1}^{n} \lambda_{j,i} \right\} \right) + r_3 \left( \exp\left\{ -r_4 \sum_{j=1}^{n} d_j \right\} \right)$$
$$+ r_5 \left( \exp\left\{ r_6 \left( \sum_{k=1}^{n} \mathcal{I}_0(d_k) - \sum_{j=1}^{m} \left(1 - \mathcal{I}_0(\tau_{j,i})\right) \right) \right\} \right) \quad (6)$$

The above equation involves three terms. The first term is a product of a sum of operating conditions ($e_{j,i}$) and an inverse exponential of the machine health conditions ($\lambda_{j,i}$) (summed over all machines). The overall value of the first term in the reward function corresponds to a decaying exponential (whose value reduces with the increase in the values of $\lambda_{j,i}$ or deterioration in the health of the machines) scaled by the summation term (whose value increases with the increase in the values of $e_{j,i}$ or decrease in favorability of the operating conditions). The implication of this first term of the reward function is that there is high reward for healthy machines which is scaled even higher under unfavorable operating conditions, i.e., it is more important to have healthy machines in unfavorable operating conditions than it is in favorable operating conditions. The second term in the reward function represents the cost of maintenance in terms of the exponential reduction in the reward due to the maintenance decisions. The third term in the reward function is an exponential term which depends on the difference between the total number of available machine and the total number of active tasks. This term is included to minimize downtime. Note that we have used an indicator function $\mathcal{I}_0(.)$. This function is defined as

$$\mathcal{I}_i(j) = \begin{cases} 1 & if\ j = i \\ 0 & if\ j \neq i \end{cases} \quad (7)$$

The exponential terms have been used in the reward function in order to ensure that the reward function is positive definite which is a favorable quality for applying stochastic dynamic programming on the MDP model for the calculation of the optimal policy. Also, $r_1, r_2, .., r_6$ are user defined constants (should be positive real numbers) to ensure adequate shaping of the reward function.

### 3.1.4 State Transition Probabilities

The details of obtaining transition probabilities for the health degradation variables and the operating condition variables are discussed in [14] and [15]. In this section, we present the state transition probability formulation in a manner that is easier to understand and comprehend. We have three types of variables in our state space. In the worst case, all of the variables show random behavior. Therefore, a generic formulation of the state transition probability is

$$P(s_j|s_i, D) = P(e_{1,j}, .., e_{n,j}, \lambda_{1,j}, ..., \lambda_{n,j}, \tau_{1,j}, ..., \tau_{m,j}|e_{1,i}, .., e_{n,i}, \lambda_{1,i}, ..., \lambda_{n,i}, \tau_{1,i}, ..., \tau_{m,i}, D) \quad (8)$$

The calculation of the above mentioned joint conditional probability can be extremely difficult in general. But fortunately, we can exploit the structure of the problem and assume conditional independence among the deterioration of the machines as well as among the operating conditions in order to simplify the state transition probabilities as

$$P(s_j|s_i, D) = \prod_{k=1}^{n} P(e_{k,j}|e_{k,i}) \prod_{k=1}^{n} P(\lambda_{k,j}|\lambda_{k,i}, e_{k,i}, D) \prod_{k=1}^{m} P(\tau_{k,j}|\tau_{k,i}) \quad (9)$$

With the above formulation, it is quite straight forward to extract the conditional probabilities from the machine data such as those discussed in [19]. Even if the data is initially unavailable, the probability values can be specified based on experience and then learnt later on as the machines are used and the data is collected.

An alternate scenario for state transitions may include machine health as an exogenous variable. As shown in [20], the life of a machine tool can be represented by a mathematical equation of the form

$$\mathcal{V}\mathcal{L}^n = \mathcal{C} \qquad (10)$$

Here, $\mathcal{V}$ is the cutting speed (in rpm), $\mathcal{L}$ is machine tool life (0 to 1), and $\mathcal{C}, n$ are appropriate constants. There are alternate formulations for the machine tool life as well (discussed in [20]) that involve cutting tool hardness, cutting temperature, feed rate etc. Therefore, it is possible to determine the machine tool life externally (by keeping track of the usage of the tool) and the value of the $\lambda$ for each machine can be provided to the MDP model at each decision-epoch. In this way, the MDP does not need to model the transition probabilities associated with the machine tool health.

## 3.2 Decision-Making Model for Machine Assignment

As discussed in the problem description, there are two decision-making modules in the setup of the problem. We have already discussed the proposed model for the machine maintenance. Here we describe the proposed model for the machine assignment to active tasks.

### 3.2.1 State Variables

The state variables for machine assignment include the information regarding the active and inactive tasks as well as the associated priority and performance specifications. We also need to know about the availability of the machines, i.e., which of the machines are currently under maintenance and which ones are not. Finally, we need the health status of the machines in order to determine the best possible performance that is deliverable by each machine. Consequently, the set of states is depicted as

$$\begin{aligned}
\mathcal{S} &= \{s_1, s_2, \ldots, s_\mathcal{N}\}, \\
s_i &= \{\lambda_{1,i}, \lambda_{2,i}, \ldots, \lambda_{n,i}, \bar{\tau}_{1,i}, \bar{\tau}_{2,i}, \ldots, \bar{\tau}_{m,i}, d_{1,i}, d_{2,i}, \ldots, d_{n,i}\}, \\
\lambda_{j,i} &\in \Lambda = \{1, 2, \ldots, L, F\}, \\
\bar{\tau}_{j,i} &\in \overline{T} = \{0, 1, 2, \ldots, p\}, \\
d_{j,i} &\in \{0, 1, 2, 3\}.
\end{aligned} \qquad (11)$$

Note that in the above state space, the degradation variables are the same as in the decision-making model for maintenance. The task variables ($\bar{\tau}_{j,i}$) in the machine assignment are different from those in the maintenance. This is because the machine assignment requires more detailed information regarding the active tasks as compared to the maintenance. In our formulation, we allow $\bar{\tau}_{j,i}$ to have $p + 1$ possible values where $\bar{\tau}_{j,i} = 0$ refers to the task being inactive whereas $p$ different values encode the combinations of possible performance and priority specifications, i.e., each value of $\bar{\tau}_{j,i}$ corresponds to a particular combination of performance and priority combination. We make the programming of the decision-making easier by assigning the performance and priority in ascending order to the variables $\bar{\tau}_{j,i}$, e.g., $\bar{\tau}_{j,i} = p$ refers to the highest possible requirements in terms of performance and priority of the task. Finally, the machine maintenance variables $d_{j,i}$ are basically the decision variables from the maintenance model. The maintenance policy conveys its decisions to the machine assignment model as shown in Figure 2 earlier. Note that if a task is assigned to a machine that is under maintenance, the execution of the task shall not begin until the machine is available again (after completion of the maintenance activity).

### 3.2.2 Set of Decisions

The set of decisions for the machine assignment model includes $n$ machine assignment variables indicating whether a machine has been assigned to an active task or not. Also, in case of an assignment, the decision must also specify which task has been assigned to which of the machines. Consequently, the mathematical representation of the set of decisions is as follows

$$A = \{a_1, a_2, \ldots, a_n\},$$
$$a_k \in \{0, 1, 2, \ldots, m\},$$
$$a_i \neq a_j, \forall (i \neq j) \& (a_i, a_j \neq 0). \tag{12}$$

The values of the variables $a_k$ in above formulation corresponds to the task assignment for the $k^{th}$ machine. Specifically, $a_k$ can assume $m + 1$ possible values where $a_k = 0$ corresponds to $k^{th}$ machine being unassigned and $a_k = j$ corresponds to $k^{th}$ machine being assigned to $j^{th}$ task. The above formulation of the decisions ensures that no two machines are assigned to the same task, i.e., $a_i \neq a_j, \forall (i \neq j)$ unless $a_i = a_j = 0$. The total number of possible decisions in our formulation is $m(m^{n-1} + n - 1) + 1$.

### 3.2.3 Reward Function

There are two objectives of the machine assignment decision-making. The first one is to achieve the desirable performance for all tasks and the second one is to incorporate task priority. Consequently, the reward function is formalized as follows

$$\mathcal{R}(s_i, A) = \rho_1 \left( \exp \left\{ -\rho_2 \sum_{j=1}^{m} \bar{\tau}_{j,i} \right\} \right) + \rho_3 \exp \left\{ -\rho_4 \left( \sum_{j=1}^{n} f(\lambda_{j,i}) \left( 1 - \mathcal{I}_0(a_j) \right) \right) \right\} \tag{13}$$

The above reward function is composed of two major terms. The first term represents the reward for completion of the tasks (note that the first term will have highest value when all of the tasks will have inactive or completed status, i.e., $\bar{\tau}_{j,i} = 0, \forall j \in \{1, 2, \ldots, m\}$). The constants $\rho_1, \ldots, \rho_4$ can be set by the user based on project-based preferences. The second term of the reward function is also an inverse exponential whose value decreases based on the function $f(\lambda_{j,i})$. This function is defined as follows

$$f(x) = \begin{cases} 1 & if\ x > \eta_1 \\ 0 & otherwise \end{cases} \tag{14}$$

The constant $\eta_1$ can be chosen based on the user preference. Recall that the values of the machine health variables $(\lambda_{j,i})$ are assigned in such a way that high values correspond to low health. Similarly, the values of the task variables $(\bar{\tau}_{j,i})$ are assigned in such a way that higher values correspond to higher priority and performance requirements. With the above-mentioned formulation, the second term penalizes the assignment of low performing machines to the tasks.

### 3.2.4 State Transition Probabilities

The state transition probabilities have two types of random variables, i.e., the machine health variables and the task assignment variables. Following the similar assumption as for the maintenance decision-making model, the simplified state transition probabilities are formulated as

$$P(s_j | s_i, A) = \prod_{k=1}^{n} P(\lambda_{k,j} | \lambda_{k,i}, d_{k,i}) \prod_{k=1}^{m} P(\bar{\tau}_{k,j} | \bar{\tau}_{k,i}) \tag{15}$$

Note that the conditional probabilities of machine health variables are common in both decision-making models. Also, the conditional probabilities for $\bar{\tau}_{k,i}$ and those of $\tau_{k,i}$ in the maintenance decision-making model are generated from the same data. Therefore, no additional data or probability distribution functions are required for the machine assignment decision-making. This is an indicator of how well the two decision-making modules within our proposed solution are integrated. Finally, note that (as discussed earlier), the machine health variable can be treated as an exogenous variable in which case its probability distribution shall not be required.

## 3.3 Calculation of the Optimal Policy and Closed-Loop Execution

We have formulated two MDP-based decision-making models in this section so far. As discussed earlier in Section 2, the MDP models are solved using stochastic dynamic programming-based approaches such as value iteration. The solution of an MDP model is an optimal policy that maps each state of the MDP with an optimal decision. The process of calculation of the optimal policy has been demonstrated in Figure *3*. The set of equations involved in the calculation of the optimal policy have been discussed earlier in the background section. Note that we need to calculate the optimal policies for both the maintenance and the machine assignment MDP models. Once the optimal policies have been computed, we can use them in the practical implementation.

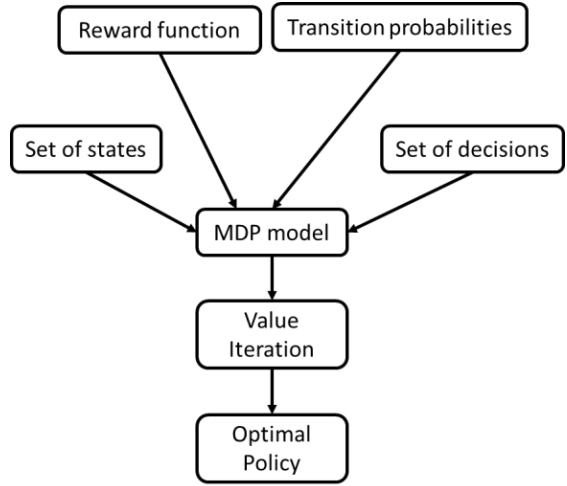

Figure 3: Steps for the calculation of an MDP-based decision-making policy

The closed loop implementation of the calculated decision-making policies has been depicted in Figure *4* where the optimal policy for machine assignment is represented by $\pi_\mathcal{A}$ and the optimal policy for machine maintenance is represented by $\pi_\mathcal{M}$. Note that time is omitted from the presentation in Figure *4*. It is implied that the flow of information is from top to bottom, i.e., in the very first decision cycle, the assignment decision shall be made before the maintenance decision. In order to facilitate this, it may be assumed that all of the machines are new, i.e., none of them is under maintenance. However, this assumption is not necessary. In principle, the values of the machine maintenance status flags ($d_{j,i}, \forall j$) can be initialize in any manner.

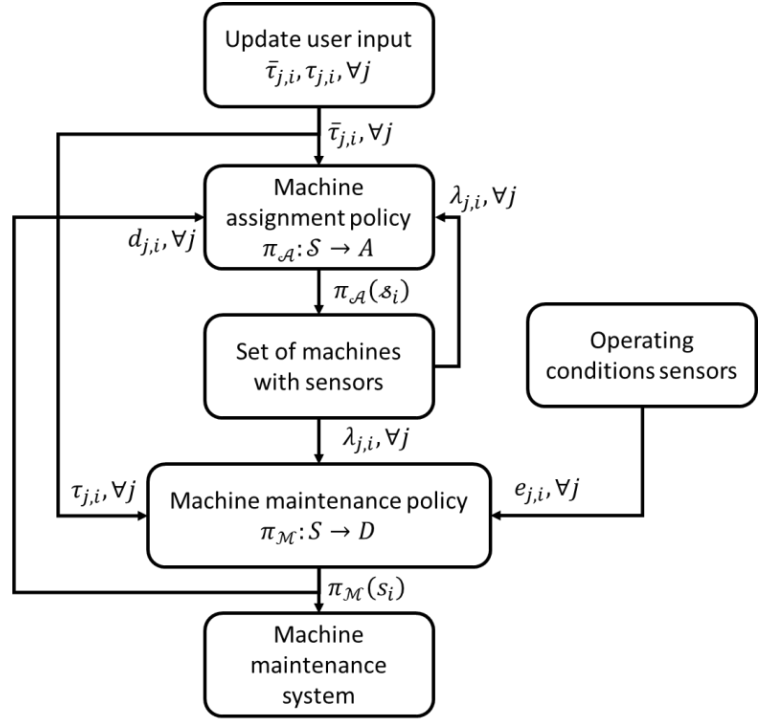

Figure 4: Closed loop implementation of the optimal policies

# 4 Case Study

This section presents a case study in order to analyze the proposed framework. In order to extract the state transition probabilities, we used the milling machine dataset [20]. The details of the data can be accessed through the reference [20]. In this section, we explain how we used the existing data to extract the state transition probabilities. Furthermore, we present analysis of the results obtained from the optimal policy calculated using value iteration algorithm.

## 4.1 Extraction of State Transition probabilities

The machine tool wear has been recorded in [20] using 16 experiments under four different operating conditions (the operating conditions in the provided dataset relate to the material that is being cut and the depth of cut requirements). Based on the given data, first of all, we calculated the rate of wear of the machine tool using simple Euler method

$$WR = \frac{W_{tf} - W_{t0}}{tf - t0} \quad (16)$$

Here, $WR$ stands for wear rate of the machine tool, $W_{tf}$ and $W_{t0}$ are the wear of the machine tool at final time ($tf$) and initial time ($t0$) respectively. Note that the wear rate turned out to be different for each operating condition. Next, we scaled the wear rate so that it may be converted from *per unit time* to *per decision epoch* of the Markov decision process. In our study, we assumed a decision epoch of 10 second but in general it can be selected as per the choice of the management. A smaller epoch results in faster response to the real-time situations but it also cost a lot of recalculations. A longer epoch on the other hand may cause delays in the response of the task assignment and health management systems. In general, we recommend the epoch time to be close to the time taken by a machine to complete one task. The values of the degradation probabilities have been listed in Table *1*. Notice that, the probabilities show that the degradation is slightly more likely when the tool is new. This is because of the inequality in the chosen mapping between the percentage health of the machine tool and the corresponding value of $\lambda$. Initially, the value of $\lambda$ changes for every 7 percent degradation but after $\lambda$ equal 3 (14% degradation), the value of $\lambda$ changes after 10% degradation. Figure *5* shows the wear in the machine tool across 16 experiments for which the data has been presented in [20]. Notice that the wear rate varies across all of the experiments. We have somewhat simplified this data in a sense that we have taken five different wear rates within the range of wear rates that can be observed from the slopes of the graphs on the left side of Figure *5*.

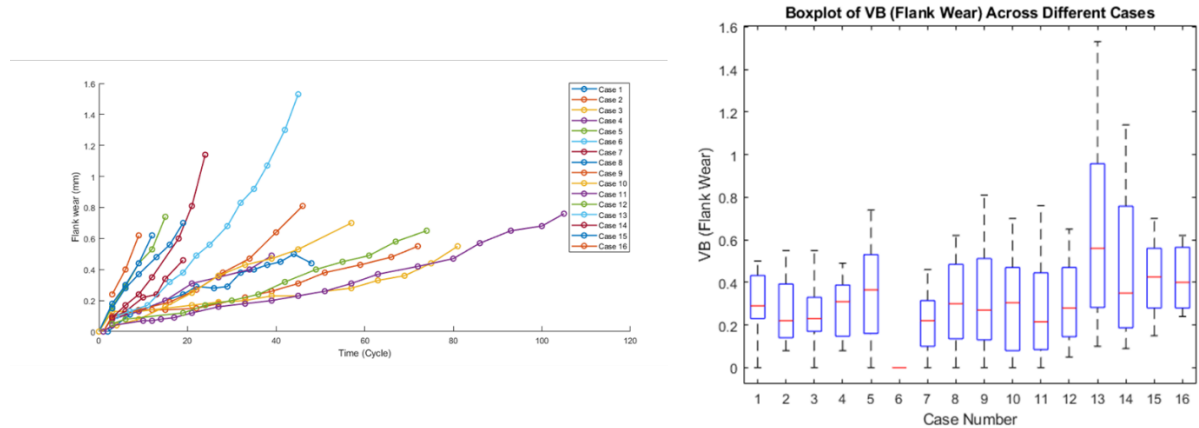

Figure 5: Visualization of the machine tool wear data in [20]

Table 1: Machine tool health degradation probabilities after scaling and rounding off

| Operating Condition ($e_i$) | Current Health ($\lambda_i$) | Health Degradation Probability $P(\lambda_j|\lambda_i, e_i)$ |
|---|---|---|
| 1 | $\lambda_i \leq 3$ (new tool) | 0.1 |
| 2 | $\lambda_i \leq 3$ (new tool) | 0.2 |
| 3 | $\lambda_i \leq 3$ (new tool) | 0.3 |
| 4 | $\lambda_i \leq 3$ (new tool) | 0.4 |
| 1 | $\lambda_i > 3$ (old tool) | 0.2 |
| 2 | $\lambda_i > 3$ (old tool) | 0.3 |
| 3 | $\lambda_i > 3$ (old tool) | 0.4 |
| 4 | $\lambda_i > 3$ (old tool) | 0.5 |

### 4.2 Setup of the Value Iteration Algorithm for Solving the MDPs

In order to obtain the optimal policies for machine task assignment and machine health management, we need to specify the parameters to be used in the value iteration algorithm. The summary of parameter values is presented in Table *2*. Note that the parameters chosen here are based on the intuition and experience of the authors and these values are only for the demonstration purposes. In general, the parameter values come from estimates of the real cost associated with machine maintenance and loss of business due to delay in task completion.

Table 2: Parameter value for the calculation of optimal policies

| Simulation parameter | Value | Remarks |
|---|---|---|
| $\gamma$ (discount factor) | 0.95 | For far-sighted policy |
| Iteration termination | $\|V^{t+1} - V^t\| < 10^{-6}$ | To ensure convergence of the value function |
| $N$ | 15,552 | number of states in the health management MDP |
| $\mathcal{N}$ | 36,864 | number of states in the machine assignment MDP |
| $L$ | 5 | Number of possible health conditions for a machine |
| $M$ | 4 | Number of possible operating conditions |
| $n, m$ | 2,3 | Number of machines, Number of tasks |
| $r_1, r_2, r_3, r_4, r_5, r_6$ | 7,0.5,10,5,5,0.25 | Cost function parameters for machine health management |
| $\rho_1, \rho_2, \rho_3, \rho_4, \eta_1$ | 100,0.5,100,2,3 | Cost function parameters for machine task assignment |

## 4.3 Trends in the Optimal Policies

One of the major advantages of using an MDP model for a decision-making problem is that we can calculate (and hence analyze) the optimal policy beforehand. In this section, we present the results obtain from the parameter settings discussed above. Figure *6* shows the histogram of different policy decisions for the machine health management. Labels on the x-axis indicate the level of maintenance applied to the machines, e.g., 1,4 means that minor maintenance is applied to the machine 1 and major maintenance is carried out for the machine 2. Notice that the occurrence of minor maintenance is less in the optimal policy as compared to the occurrence of major maintenance of the machines. This is an indication of the policy trying to delay the scheduling of the maintenance in favor of maximizing the machine availability for the tasks. Figure *7* shows the results obtained from the task assignment policy. In this case, the x-axis labels indicate the index of the task assigned to a machine, e.g., 3,1 means that the machine 1 is assigned to the task 3 and the machine 2 is assigned to the task 1. Notice that the assignment of task 3 is more popular in the policy because it in order to break the tie amongst the tasks, the policy selects the task with a higher index. For example, if only one machine is available and all three tasks are pending with equal priority, the machine will be assigned to the task with highest index. Another noticeable trend in Figure *7* is that the policy prefers to keep both machines assigned rather than assigning only one of the machines.

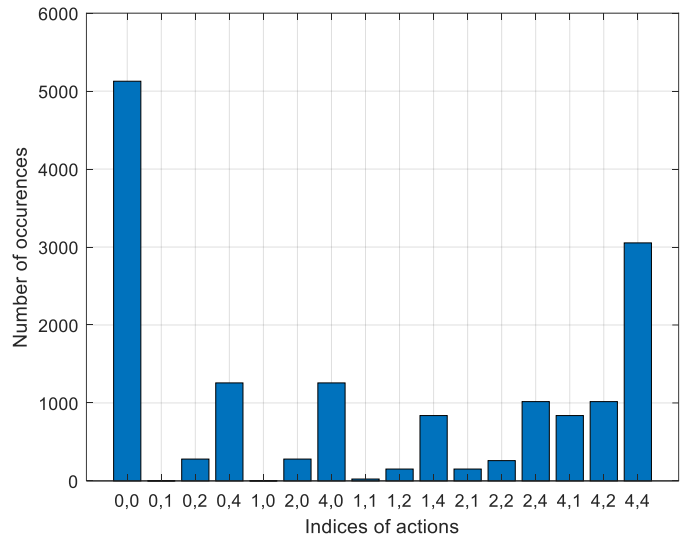

Figure 6: Trends in the optimal policy for machine health management

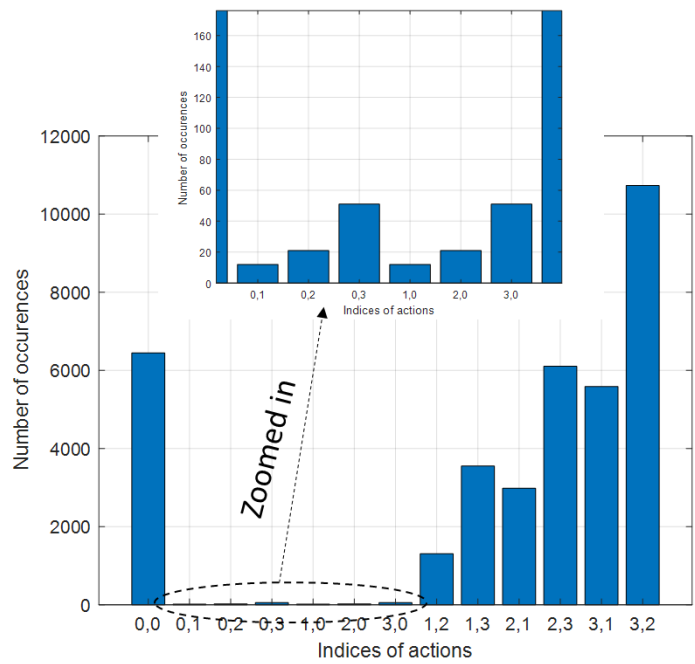

Figure 7: Trends in the machine task assignment policy

## 4.4 Sample Trajectories

Now we discuss some sample trajectories obtained by using the policies. Note that for the trajectories presented in this section, we decide the state transition based on the maximum likelihood, i.e., given the execution of an action from a state, the next state (from all possible next states) is chosen based on the maximum probability of occurrence. Further information regarding the generation of the trajectories is presented in Table *3*. Notice that, both trajectories have been generated for 7 decision epochs (based on the fact that this much length is enough to deal with any active tasks and required maintenance. The results from the trajectory 1 are shown in Figure *8* and Figure *9*. Since the initial health of the machines is not good in trajectory 1, the maintenance decision is executed during the first decision epoch. As a result, the machines are not available during the second decision epoch ($d_1$ and $d_2$ become equal to 2). But before undergoing the maintenance, tasks 2 and 3 have been completed by the machine in the duration between epoch 1 and epoch 2. After going through the maintenance (during the time between epoch 2 and epoch 3), the machines are assigned the tasks 3 and 1. Since task 3 has already been completed, the assignment of a machine to this task means that the machine shall be waiting for the task to reactivate. Task 1 on the other hand gets completed by the fourth decision epoch. Overall, the policies of machine health management and task assignment are able to perform well during this trajectory.

Trajectory 2 is designed to pose extra challenge for the decision-making policies by intervention of degradation in machine health and reactivation of tasks at the fourth decision epoch. The results are shown in Figure *10* and Figure *11*. Notice that initially, no maintenance is needed so the active tasks are completed within three decision epochs in the order of their respective priority. Then, once the intervention is introduced at the fourth decision epoch, the maintenance policy puts the machines under maintenance at the fifth decision epoch and the reactivated tasks are completed again by the eighth decision epoch. Note that the eighth decision epoch is only representing the transition of the state resulting from the seventh decision (we only make seven decisions per trajectory as mentioned in Table *3*).

The results from the sample trajectories clearly indicate the ability of our proposed decision-making framework to tackle the machine maintenance and the completion of the tasks while accounting for the task priorities.

Table 3: Trajectory related information

|  | Trajectory 1 | Trajectory 2 |
| --- | --- | --- |
| Initial State | $e_1 = 3, e_2 = 3, l_1 = 3,$ $l_2 = 3, d_1 = 0, d_2 = 0,$ $t_1 = 1, t_2 = 2, t_3 = 2$ | $e_1 = 3, e_2 = 3, l_1 = 1,$ $l_2 = 1, d_1 = 0, d_2 = 0,$ $t_1 = 1, t_2 = 2, t_3 = 2$ |
| Number of decision epochs | 7 | 7 |
| Interventions | *none* | Change the states to $e_1 = 3, e_2 = 3, l_1 = 3,$ $l_2 = 3, d_1 = 0, d_2 = 0,$ $t_1 = 1, t_2 = 2, t_3 = 2$ at fourth decision epoch (after the fourth decision) |

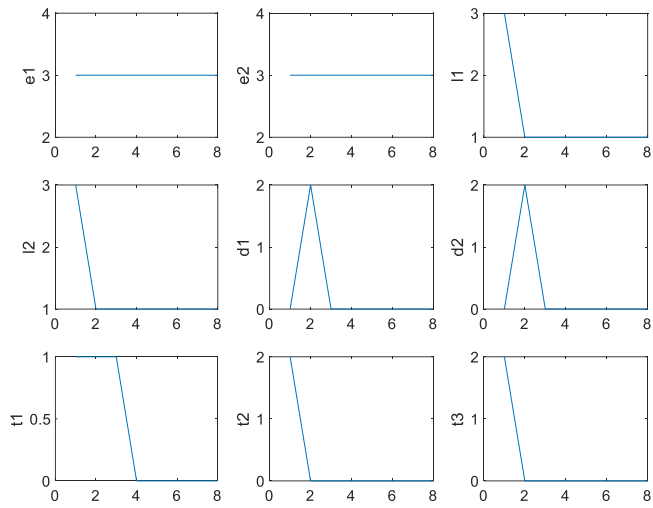

Figure 8: State variable values for trajectory 1

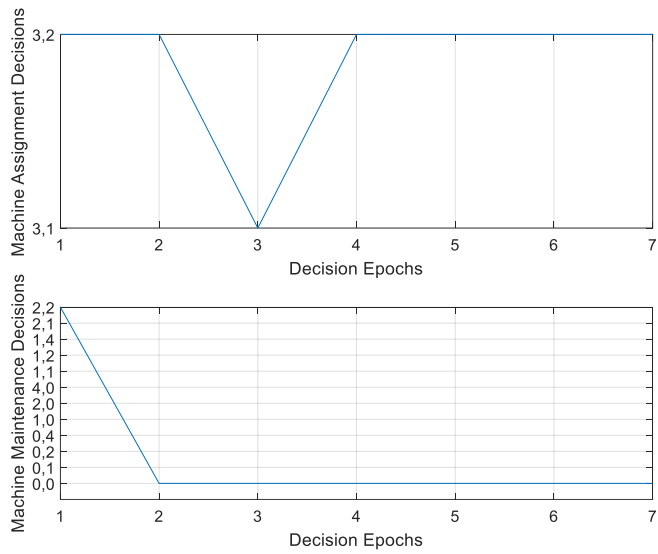

Figure 9: Decisions taken during trajectory 1

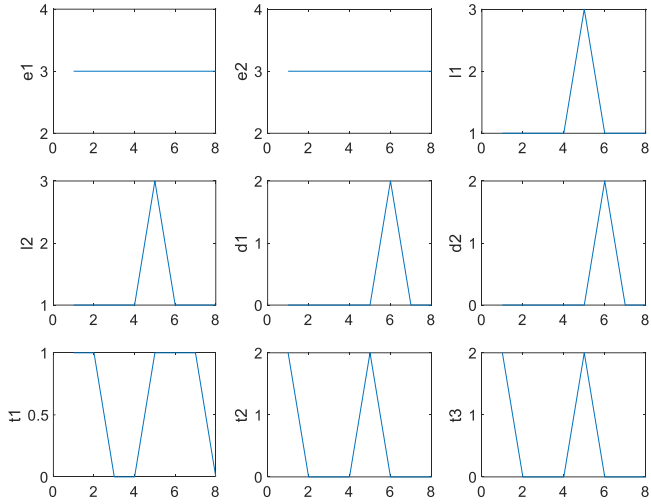

Figure 10: State variable values for trajectory 2

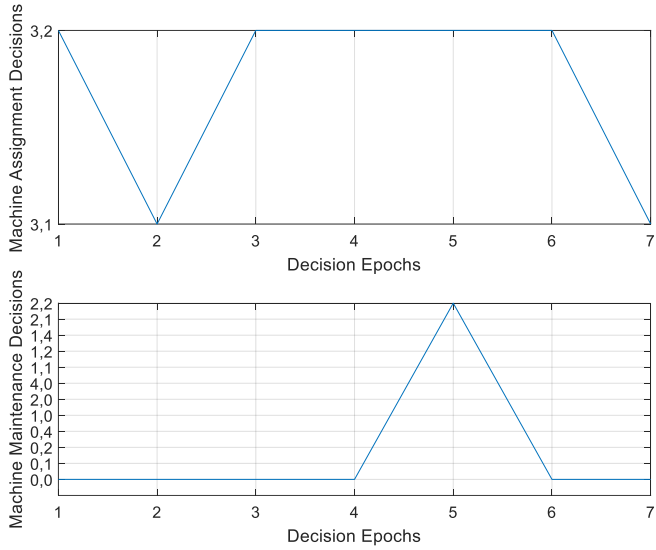

Figure 11: Decisions taken during trajectory 2

# 5   Conclusions

We have discussed a new MDP-based framework for simultaneous machine health management and task assignment for the machines under variable and uncertain operating conditions. The framework allows for offline calculation of the optimal decision-making policies which in turn greatly increases the online response time. Furthermore, the parameter values for the framework can be optimized by reverse engineering the optimal policy (in situations where the actual costs related to the missing of task deadlines or machine maintenance are unknown). The proposed framework offers modeling of uncertainty and collaboration between the machine health maintenance and the machine utilization decision-making. We believe that the proposed framework can serve as a base for more optimized and highly scalable solutions for the problem discussed in this paper. Specifically, one of the possible future avenues of research is to use machine learning for parameter estimation and optimization and hence periodic recalculation of the optimal decision-making policies. Another possible future direction is to investigate distributed task assignment and health management decision-making for large scale problems having many machines and many tasks.

**Declaration of interests**

The authors declare that they have no known competing financial interests or personal relationships that could have appeared to influence the work reported in this paper.

**Acknowledgement**

The authors would like to acknowledge the support provided by the Deanship of Research Oversight and Coordination (DROC) at King Fahd University of Petroleum & Minerals (KFUPM) for funding this work through IRC-IMR grant number INMR2301.

## References

[1] Nunes, P., J. Santos, and E. Rocha. "Challenges in predictive maintenance–A review." *CIRP Journal of Manufacturing Science and Technology* 40 (2023): 53-67.
[2] Ahmad, Rosmaini, and Shahrul Kamaruddin. "An overview of time-based and condition-based maintenance in industrial application." *Computers & industrial engineering* 63.1 (2012): 135-149.
[3] Prajapati, Ashok, James Bechtel, and Subramaniam Ganesan. "Condition based maintenance: a survey." *Journal of Quality in Maintenance Engineering* 18.4 (2012): 384-400.
[4] Alaswad, Suzan, and Yisha Xiang. "A review on condition-based maintenance optimization models for stochastically deteriorating system." *Reliability engineering & system safety* 157 (2017): 54-63.
[5] Li, Y., S. Peng, Y. Li, and W. Jiang. "A review of condition-based maintenance: Its prognostic and operational aspects." *Frontiers of Engineering Management* 7.3 (2020): 323-334.
[6] Quatrini, E., F. Costantino, G.D. Gravio, and R. Patriarca. "Condition-based maintenance—an extensive literature review." *Machines* 8.2 (2020): 31.
[7] Ali, Ahad, and Abdelhakim Abdelhadi. "Condition-based monitoring and maintenance: state of the art review." *Applied Sciences* 12.2 (2022): 688.
[8] Nebelung, Nicolas, et al. "Towards Real-Time Machining Tool Failure Forecast Approach for Smart Manufacturing Systems." *IFAC-PapersOnLine* 55.2 (2022): 548-553.
[9] Liao, Linxia, and Jay Lee. "Design of a reconfigurable prognostics platform for machine tools." *Expert systems with applications* 37.1 (2010): 240-252.
[10] Liu, C., L. Zhang, J. Niu, R. Yao, and C. Wu. "Intelligent prognostics of machining tools based on adaptive variational mode decomposition and deep learning method with attention mechanism." *Neurocomputing* 417 (2020): 239-254.
[11] Hu, K., Y. Cheng, J. Wu, H. Zhu, X. Shao. "Deep bidirectional recurrent neural networks ensemble for remaining useful life prediction of aircraft engine." *IEEE Transactions on Cybernetics* (2021).
[12] Liu, Q., W. Liu, M. Dong, Z. Li, and Y. Zheng. "Residual Useful Life Prognosis of Equipment Based on Modified Hidden semi-Markov Model with a Co-evolutional Optimization Method." *Computers & Industrial Engineering* (2023): 109433.
[13] Zheng, P., W. Zhao, Y. Lv, L. Qian, and Y. Li. "Health status-based predictive maintenance decision-making via LSTM and Markov decision process." *Mathematics* 11.1 (2022): 109.
[14] Diyin, T. A. N. G., C. A. O. Jinrong, and Y. U. Jinsong. "Remaining useful life prediction for engineering systems under dynamic operational conditions: A semi-Markov decision process-based approach." *Chinese Journal of Aeronautics* 32.3 (2019): 627-638.
[15] Long, J., C. Chen, Z. Liu, J. Guo, and W. Chen. "Stochastic hybrid system approach to task-orientated remaining useful life prediction under time-varying operating conditions." *Reliability Engineering & System Safety* 225 (2022): 108568.
[16] Vachtsevanos, G. J., F. Lewis, M. Roemer, A. Hess, and B. Wu. *Intelligent fault diagnosis and prognosis for engineering systems*. Vol. 456. Hoboken: Wiley, 2006.

[17] Puterman, Martin L. *Markov decision processes: discrete stochastic dynamic programming*. John Wiley & Sons, 2014.

[18] Powell, Warren B. *Approximate Dynamic Programming: Solving the curses of dimensionality*. Vol. 703. John Wiley & Sons, 2007.

[19] Eker, Omer Faruk, Faith Camci, and Ian K. Jennions. "Major challenges in prognostics: Study on benchmarking prognostics datasets." *PHM Society European Conference*. Vol. 1. No. 1. 2012.

[20] Agogino, A., and Goebel, K. "Mill Data Set", BEST lab, UC Berkeley. NASA Ames Prognostics Data Repository, [http://ti.arc.nasa.gov/project/prognostic-data-repository], NASA Ames, Moffett Field, CA, 2007.